\documentclass[journal]{IEEEtran}

\usepackage{cite}

\usepackage{color}

\usepackage{amsmath}

\usepackage{algorithm}
\usepackage{algorithmic}
\usepackage{bm}
\usepackage{latexsym}
\usepackage{amsthm}
\usepackage{url}
\usepackage{amsfonts}
\usepackage{amssymb}
\usepackage{indentfirst}
\usepackage{float}
\ifCLASSINFOpdf
   \usepackage[pdftex]{graphicx}
   \graphicspath{{../pdf/}{../jpeg/}}
   \DeclareGraphicsExtensions{.pdf,.jpeg,.png}
\else
   \usepackage[dvips]{graphicx}
   \graphicspath{{../eps/}}
   \DeclareGraphicsExtensions{.eps}
\fi
\usepackage{array}
\usepackage{multirow}

\ifCLASSOPTIONcompsoc
  \usepackage[caption=false,font=normalsize,labelfont=sf,textfont=sf]{subfig}
\else
  \usepackage[caption=false,font=footnotesize]{subfig}
\fi

\usepackage{setspace}
\begin{document}

\title{ {Perceptive self-supervised learning network for noisy image watermark removal}}

\author{Chunwei Tian, \emph{Member}, \emph{IEEE},
        Menghua Zheng,
        Bo Li, 
        Yanning Zhang, \emph{Senior Member}, \emph{IEEE},
        Shichao Zhang, \emph{Senior Member}, \emph{IEEE},
        David Zhang, \emph{Life Fellow}, \emph{IEEE}
        % <-this % stops a space
\thanks{This work was supported in part by the National Natural Science Foundation of China under Grant 62201468, in part by the China Postdoctoral Science Foundation under Grant 2022TQ0259 and 2022M722599, in part by the Youth Science and Technology Talent Promotion Project of Jiangsu Association for Science and Technology under Grant JSTJ-2023-017. (Corresponding author: Bo Li (Email: libo803@nwpu.edu.cn), Yanning Zhang (Email: ynzhang@nwpu.edu.cn)) }
\thanks{Chunwei Tian is with School of Software, Northwestern Polytechnical University, Xi’an, 710129, China. Also, he is with National Engineering Laboratory for Integrated Aero-Space-Ground-Ocean Big Data Application Technology, Xi’an, 710129, China. (Email: chunweitian@nwpu.edu.cn)}
\thanks{Menghua Zheng is with School of Software, Northwestern Polytechnical University, Xi’an, 
710129, China. (Email: menghuazheng@mail.nwpu.edu.cn)}
\thanks{Bo Li is with School of Electronics and Information, Northwestern Polytechnical University, Xi’an, 710129, China. (Email: libo803@nwpu.edu.cn)}
\thanks{Yanning Zhang is with School of Computer Science, Northwestern Polytechnical University, National Engineering Laboratory for Integrated Aero-Space-Ground-Ocean Big Data Application Technology, Xi’an, 710129, China. (Email: ynzhang@nwpu.edu.cn)}
\thanks{Shichao Zhang is with Guangxi Key Lab of Multisource Information Mining \& Security College of Computer Science \& Engineering Guangxi Normal University, Guilin, 541004, China. (Email: zhangsc@mailbox.gxnu.edu.cn)}
% School of Computer Science and Engineering, Central South University, Changsha, 410083, China. 
\thanks{David Zhang is with School of Data Science, The Chinese University of Hong Kong (Shenzhen), Shenzhen, 518172, China. Also, he is with Shenzhen Institute of Artificial Intelligence and Robotics for Society, Shenzhen, 518172, China. (Email: davidzhang@cuhk.edu.cn)}
}

%\markboth{~}%
%{Shell \MakeLowercase{\textit{et al.}}: Bare Demo of IEEEtran.cls for Journals}

\maketitle

\begin{abstract}
Popular methods usually use a degradation model in a supervised way to learn a watermark removal model. However, it is true that reference images are difficult to obtain in the real world, as well as collected images by cameras suffer from noise. To overcome these drawbacks, we propose a perceptive self-supervised learning network for noisy image watermark removal (PSLNet) in this paper. PSLNet depends on a parallel network to remove noise and watermarks. The upper network uses task decomposition ideas to remove noise and watermarks in sequence. The lower network utilizes the degradation model idea to simultaneously remove noise and watermarks. Specifically, mentioned paired watermark images are obtained in a self-supervised way, and paired noisy images (i.e., noisy and reference images) are obtained in a supervised way. To enhance the clarity of obtained images, interacting two sub-networks and fusing obtained clean images are used to improve the effects of image watermark removal in terms of structural information and pixel enhancement. Taking into texture information account, a mixed loss uses obtained images and features to achieve a robust model of noisy image watermark removal. Comprehensive experiments show that our proposed method is very effective in comparison with popular convolutional neural networks (CNNs) for noisy image watermark removal.  Codes can be obtained at https://github.com/hellloxiaotian/PSLNet.

\end{abstract}


% \IEEEpeerreviewmaketitle
\section{Introduction}
With the improvements in digital devices, images have played more important roles 
between human-computer interactions \cite{clouard2011human}. To protect ownership of these images, watermark removal techniques are presented \cite{hu2005algorithm}\cite{hu2006reversible}. Although these watermarks have positive effects on protecting ownership, they may suffer from some challenges of watermark techniques \cite{chen2021refit}. To verify the robustness of these watermarks, image watermark removal techniques are developed \cite{qin2018visible}. For instance, Westfeld et al. \cite{westfeld2008regression} used locally relational parts to predict images for estimating watermarks. To improve the effects of image watermark removal, Hsu et al. \cite{hsu2011new} exploited stochastic models to establish relations among given images, added watermarks, and watermarked images to reduce irrelevant information for image watermark removal. To verify the robustness of added watermarks in images and videos, singular value decomposition was used to delete watermarks \cite{nikbakht2015targeted}. To automatically remove watermarks, Xu et al. exploited a thresholding algorithm to detect watermark areas and remove watermarks in obtained areas \cite{xu2017automatic}. To make a tradeoff between watermark removal performance and efficiency, the total variation method utilized edges of obtained structure images from given watermark images to detect watermarks and remove them \cite{santoyo2017automatic}. Although these methods have performed well in image watermark removal, they face a shortage of manual setting parameters and complex optimization methods.

To overcome these drawbacks, deep learning techniques with deep network architectures are conducted for low-level vision \cite{tian2022generative}\cite{tian2022heterogeneous}. Yue et al. combined three different networks and patch matching strategies (i.e., global and local ways) to remove noise when targeted images suffered from large deformation \cite{yue2019ienet}. Due to strong learning abilities in deep convolutional neural networks (CNNs), deep CNNs have been widely used for image watermark removal \cite{cheng2018large}. To deal with watermark removal with watermarks at random locations and different angles, a generative adversarial network (GAN) with a self-attention mechanism was used to suppress these special watermarks \cite{cao2019generative}. To cope with complex watermarks, i.e., opaque and semi-transparent watermarks, a conditional GAN was utilized to keep more detailed information when watermarks were erased \cite{li2019towards}. Alternatively, a two-stage network was designed to extract watermarks and repair images to prevent the negative effects of backgrounds on watermark removal \cite{jiang2020two}. To recover more detailed information, a combination of residual UNets and attention mechanisms was used to remove watermarks by different stages. Also, a perceptual loss was referred to recover more texture information to improve the effects of image watermark removal \cite{cun2021split}. Similarly, Liu et al. \cite{liu2021wdnet} implemented a first phase network to find watermark regions and used the second phase network to remove watermarks from these regions to overcome limitations, i.e., shapes, transparency, sizes, and color. To improve the visual effects of watermark removal, a serial architecture composed of two stacked U-Nets was utilized to facilitate salient hierarchical information to effectively remove watermarks \cite{fu2022improved}. Federated learning can extract high-level feature information to better reconstruct clean images \cite{li2022image}. Additionally, watermark vaccine methods can also better remove watermarks \cite{liu2022watermark}. Although these methods are very effective in removing watermarks, they rely on reference images in a supervised manner to learn watermark removal models. However, 
reference images are not easy to obtain in the real world. Besides, collected images often suffer from noise interference. 

    \begin{figure*}[!htbp]
\centering
\subfloat{\includegraphics[width=7in]{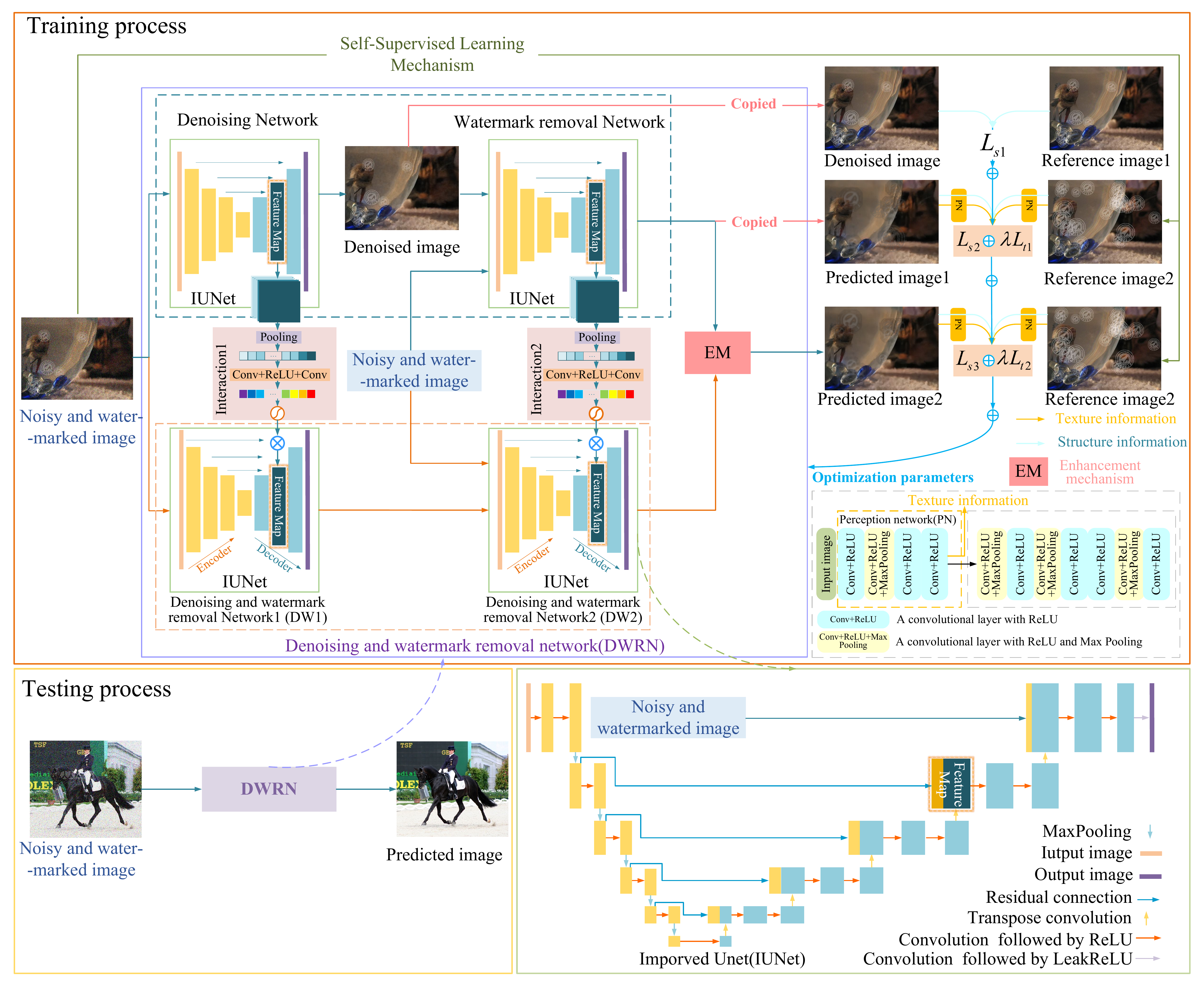}
  }
\caption{Architecture of PSLNet. It contains training and test processes. The training process uses four IUNet for image denoising and watermark removal. A preception network is used to extract more texture information for image watermark removal.} 
\vspace{-2.0em}
\end{figure*}

In this paper, we present a self-supervised learning network for noisy image watermark removal as well as PSLNet. PSLNet uses a parallel network to eliminate watermarks in given noisy images. The upper network exploits the task decomposition idea to remove noise and watermarks in sequence. The lower network utilizes the degradation model idea to simultaneously remove noise and watermarks. Specifically, mentioned paired watermark images are obtained in a self-supervised way, and paired noisy images (i.e., noisy and reference images) are obtained in a supervised way. To obtain more clearer images, interacting two sub-networks and fusing obtained clean images from two sub-networks can improve visual effects in terms of structural information and pixel enhancement for image watermark removal. Also, a mixed loss is used to facilitate more perceptual information to obtain more texture information for noisy image watermark removal. Our proposed PSLNet is very competitive for noisy image watermark removal in terms of quantitative and qualitative analysis. 

The contributions of this paper can be summarized as follows. 
    
(1)	A self-supervised method is used to address image watermark removal of non-reference images. 

(2)	 The proposed method uses decomposition and degradation model ideas for 
removing watermarks in noisy images.

(3)	Enhancing structural information and pixel enhancement can improve the visual effects of noisy image watermark removal.

(4) A structural and texture loss can be used to improve the performance of noisy image watermark removal.

The remaining organizations of this paper can be presented as follows. Section II illustrates our proposed method. Section III shows our method analysis and experimental results. Section IV lists our conclusions. 

\section{Proposed method}
 {\subsection{Self-supervised method for obtaining paired watermark images}}
It is known that most image restoration methods rely on given reference images 
in a supervised way to achieve image restoration models \cite{tian2020deep}. However, reference images are not easy to obtain in the real world. Also, the image watermark removal task is faced with the same challenge. That can be addressed by only given images, which is called a self-supervised method \cite{lehtinen2018noise2noise}. That is, we assume that an input pair (the \emph{jth} input image $I_j$, the \emph{jth} label $O_j$) can be represented as  $({I_j},{O_j})$. Using an objective function to train a CNN can be summarized as Eq. (1).

 {
\begin{footnotesize}
\begin{equation}
\mathop {\arg \min }\limits_p {E_{(I,O)}}\{ D({f_p}(I),O)\}, 
\end{equation}
\end{footnotesize}}

\noindent{where $f_p$ is a CNN, $D$ is an objective function and $p$ denotes obtained parameters. And $E_{(I,O)}$ denotes the expectation of $D$ about $({I_j},{O_j})$.} 

According to Bayesian theory, the whole training task can be decomposed into several training steps. That is, Eq. (1) can be transformed as Eq. (2). 

{\begin{footnotesize}
\begin{equation}
\mathop {argmin}\limits_p {E_I}\{ {E_{O|I}}\{ D({f_p}(I),O)\} \},
\end{equation}
\end{footnotesize}}

\noindent{where $E_I$ and $E_{O|I}$ denote the expectation and conditional expectation of $D$ about the $I$ and $O$.} According to Eq. (2), we can see that the relation of input and target is one-on-many rather than one-on-one. Thus, Eq. (2) can be represented as Eq. (3).

\begin{footnotesize}
\begin{equation}
\mathop {argmin}\limits_p \sum\limits_j {D(f_{p}(I_{j}^{'}),O_{j}^{'})},
\end{equation}
\end{footnotesize}

\noindent{where input and target have the same noisy distribution and they are satisfied to $E\{ O_{j}^{'}|I_{j}^{'}\} = O_{j}^{'}$. $I_{j}^{'}$ is an input noise image. $O_{j}^{'}$ is an output noise image with the same distribution as the input noise image.

According to these illustrations, we can see that noisy images with noise of the same distribution can be treated as reference images. Motivated by that \cite{lehtinen2018noise2noise}, we use a self-supervised method to conduct paired images (i.e., watermark images and reference images), according to given watermark images. That is, a watermark image is conducted via randomly adding watermarks once on a clean image and a reference image is obtained via randomly adding watermarks on the same clean image. Watermark and reference images are used to conduct a pair of images for training a watermark removal model. Besides, watermarked images can be conducted via Eq. (4) \cite{dekel2017effectiveness}.}

\begin{footnotesize}
\begin{equation}
{I_w}(l) = \partial (l)W(l) + (1 - \partial (l)){I_c},
\end{equation}
\end{footnotesize}

\noindent{where $l(x,y)$ is used to denote pixel location. $\partial (l)$ represents spatially varying opacity. Also, ${I_w}$ and $W$ are symbolised as a watermark image and an added watermark. $I_c$ is a given clean image. More information of the process can be found in  \cite{dekel2017effectiveness}.

 {\subsection{Network architecture}}

PSLNet includes three parts, i.e., a self-supervised learning mechanism (SSL), a denoising and watermark removal network (DWRN) and a stacked perception network (SPN), which is shown in Fig. 1. SSL utilizes Section II.A to obtain pair images, i.e., watermarked image and reference image. DWRN is composed of two parallel networks to remove watermarks in noisy images. Specifically, the 36-layer upper network consists of two stacked improved U-Nets to remove noise and watermarks in sequence, according to the task decomposition principle. The first U-Net is used to remove noise. The second U-Net is responsible for removing watermarks. The 36-layer lower network includes two same stacked improved U-Nets to simultaneously remove noise and watermarks, according to the degradation model idea. It is noted that paired noisy images (i.e., noise images and reference images) are conducted via additive Gaussian noise in a supervised way, which has the same idea as Ref. \cite{zhang2017beyond}. Its process can be shown in Eq. (5).

\begin{footnotesize}
\begin{equation}
\begin{array}{ll}
O &= PSLNet(I)\\
{\rm{   }} &= {O_{Upper}}\& {O_{Lower}}\\
{\rm{   }} &= Upper(I)\& Lower(I)\\
{\rm{   }} &= (IUNet(IUNet(I)))\& IUNet(IUNet(I)),
\end{array}
\end{equation}
\end{footnotesize}

\noindent{where $PSLNet$, $Upper$, and $Lower$ are used to denote functions of PSLNet, Upper network, and Lower network, respectively. $O_{Upper}$ and $O_{Lower}$ are outputs of upper and lower networks, respectively. $\&$ stands for multiple interactions of upper and lower networks. $IUNet$ is a function of an improved U-Net \cite{ronneberger2015u}, where the input image rather than the output of the second convolutional layer} acts on the highest connection to enhance the effect of original information to improve the effects of image watermark removal.

To enhance the clarity of obtained images, multiple interactions are used between two sub-networks to improve the effects of image watermark removal. The first and second interactions depend on structural information to enhance the effects of noisy image watermark removal. That is the first interaction i.e. Interaction1 \cite{hu2018squeeze} composed of a stacked pooling, Conv, ReLU, Conv, and Sigmoid is acted between a denoising network in the upper network and denoising and watermark network1 (DW1) to suppress noise, where Conv is a convolutional layer. The second interaction i.e. Interaction2 \cite{hu2018squeeze} including a stacked pooling, Conv, ReLU, Conv, and Sigmoid is acted between a watermark removal network in the upper network and denoising and watermark network 2 (DW2) to remove watermarks. To further improve the clarity of obtained images, an enhancement mechanism containing two phases is proposed, according to pixel enhancement. The first phase uses a concatenation operation to fuse obtained two images of upper and lower networks to enhance images. To prevent over-enhancement phenomenon, a combination of a convolutional layer and a LeakyReLU is used to eliminate redundant information. The procedure can be symbolised as Eqs. (6)-(9). 

\vspace{-0.3cm}
\begin{equation}
\begin{array}{ll}
%O &= {O_{Upper}}\& {O_{Lower}}\\
O {\rm{   }} &= EM({O_{Upper}},{O_{Lower}})\\
{\rm{   }} &= EM(IUNet(IUNet(I)),{O_{Lower}}),
\end{array}
\end{equation}
\vspace{-0.6cm}

\vspace{-0.3cm}
\begin{equation}
    EM = CL(Concat(IUNet(IUNet(I)),{O_{Lower}})),
\end{equation}

\vspace{-0.6cm}

\begin{equation}
\begin{array}{ll}
    {O_{Lower}} =& DW2(Interaction1(DN(I)) \\ & \otimes DW1(I)) 
    \otimes Interaction2(WRN(I)),
\end{array}
\end{equation}
\vspace{-0.6cm}

\begin{equation}
\begin{array}{ll}
Interaction1() &= Interaction2()\\
 &= SC(RC(Pooling())),
\end{array}
\end{equation}
% \vspace{-0.6cm}

\noindent{where $EM$ denotes enhancement mechanism. $CL$ is a combination of a convolutional layer and a LeakyReLU. $Interaction1$ and $Interaction2$ denote the first interaction and the second interaction. $DN$, $DW1$, $DW2$, and $WRN$ stand for a denoising network, denoising and watermark removal network 1, denoising and watermark removal network 2, and a watermark removal network, respectively. $Pooling$, $RC$, and $SC$ are used to represent a pooling operation, a combination of a convolutional layer and a ReLU, a combination of a convolutional layer and Sigmoid, respectively.}
 
{\subsection{Perception network}}
To extract more texture information, a 16-layer perception network is used, which 
is implemented by a VGG \cite{li2019towards}. That is, a VGG is trained well on ImageNet. Then, obtained images of PSLNet and conducted reference images (watermarked images of the same distribution as input images) are acted on VGG to extract perception information. The first four layers of VGG as well as the perception network (PN) are used to monitor texture features, where the outputs of the 4th layer are used to conduct a mixed loss function. Also, the first four layers are composed of a Conv+ReLU, Conv+ReLU+MaxPooling, and two stacked Conv+ReLU, where a Conv+ReLU denotes a combination of a convolutional layer and a ReLU. A Conv+ReLU+MaxPooling denotes a combination of a convolutional layer, a ReLU, and a max-pooling operation. Its implementation can be transformed as follows. 

\begin{equation}
\begin{array}{cc}
     {f_{PN1}} = 2CR(CRM(CR({f_{vgg}})))
\end{array}
\end{equation}

\noindent{where $f_{vgg}$ is a function of VGG, $CRM$ denotes a Conv+ReLU+MaxPooling. $iCR$ denotes $i$ stacked combinations of convolutional layer and ReLU, where $i=1,2$. Also, Fig. 1 shows the architecture of PN.} The remaining network of VGG besides the PN is used to achieve a robust classifier, which can help PN improve its ability to extract texture information. 

{\subsection{Mixed loss function}}

A mixed loss based on L1 is conducted for noisy image watermark removal, according to structural and texture information. It contains two parts, i.e., structural and texture loss. Specifically, the structural loss can be used to verify robustness of PSLNet for obtaining structural information. Also, texture loss can measure robustness of PSLNet via a perception network for obtaining texture information. To accelerate the convergence, L1 loss is chosen to train a noisy watermark removal model. Besides, training images can be obtained via combining a self-supervised mechanism and a supervised mechanism in Sections II.A and II.B. Specifically, structural loss as well  $L_s$ includes three parts, i.e. $L_{s1}$ , $L_{s2}$ and $L_{s3}$. $L_{s1}$ is applied on a denoising network from the upper network to remove noise.  $L_{s2}$ is applied on a watermark removal network from the upper network to remove watermarks. $L_{s3}$ is applied on DWRN to simultaneously remove noise and watermarks. Perceptual loss i.e. $L_{t}$ consists of two parts, i.e., $\lambda L_{t1}$ and $\lambda L_{t2}$. $\lambda L_{t1}$ uses predicted images of the upper network and obtained references as inputs of PN to extract texture features to compute loss value.  $\lambda L_{t2}$ utilizes predicted images of DWRN and obtained references as inputs of PN to extract texture features to compute loss value. To make readers understand this process, the mentioned mixed loss function can be shown as follows. 

\begin{small}
\begin{equation}
    \begin{array}{ll}
L&= {L_s} + {L_t}\\
&={L_{s1}} + {L_{s2}} + {L_{s3}} + \lambda {L_{t1}} + \lambda {L_{t2}}\\
&= {\raise0.7ex\hbox{$1$} \!\mathord{\left/
 {\vphantom {1 N}}\right.\kern-\nulldelimiterspace}
\!\lower0.7ex\hbox{$N$}}\sum\limits_{j = 1}^{N} {\left| {DN(I_{nw}^{j}) -  I_{sd}^{j}} \right|}  + {\raise0.7ex\hbox{$1$} \!\mathord{\left/
 {\vphantom {1 N}}\right.\kern-\nulldelimiterspace}
\!\lower0.7ex\hbox{$N$}}\sum\limits_{j = 1}^{N} {\left| {WRN(O_{DN}^{j}) - I_s^{j}} \right|} \\& + {\raise0.7ex\hbox{$1$} \!\mathord{\left/
 {\vphantom {1 N}}\right.\kern-\nulldelimiterspace}
\!\lower0.7ex\hbox{$N$}}\sum\limits_{j = 1}^{N} {\left| {DWRN(I_{nw}^{j}) - I_{s}^{j}} \right|}  + \lambda {L_{t1}} + \lambda {L_{t2}}\\
&=  {L_{s1}} + {L_{s2}} + {L_{s3}} + \lambda \left| {{f_{PN}}(O_{Upper}^{j}) - {f_{PN}}(I_{s}^{j})} \right| + \\& \lambda \left| {{f_{PN}}(O_{DWRN}^{j}) - {f_{PN}}(I_{s}^{j})} \right|,
\end{array}
\end{equation}
\end{small}

\noindent{where $N$ stands for the number of all noisy watermarked images. $\lambda$ denotes an adjustment coefficient for obtaining texture information. $I_{nw}^{j}$ and $I_{sd}^{j}$ are denoted as the \emph{jth} watermark image with noise and obtained reference watermark image without noise, respectively. $O_{DN}^{j}$ and  $I_{s}^{j}$ are used to represent the \emph{jth} output image of DN and reference image of the DWRN. $O_{Upper}^{j}$ and $O_{DWRN}^{j}$ are \emph{jth} output images of upper networks and DWRN, respectively. Besides, the parameters of our noisy image watermark removal model can be optimized by Adam \cite{kingma2014adam}. 
}

\section{Experiments}
\begin{figure}[!htbp]
\centering
\subfloat{\includegraphics[width=3in]{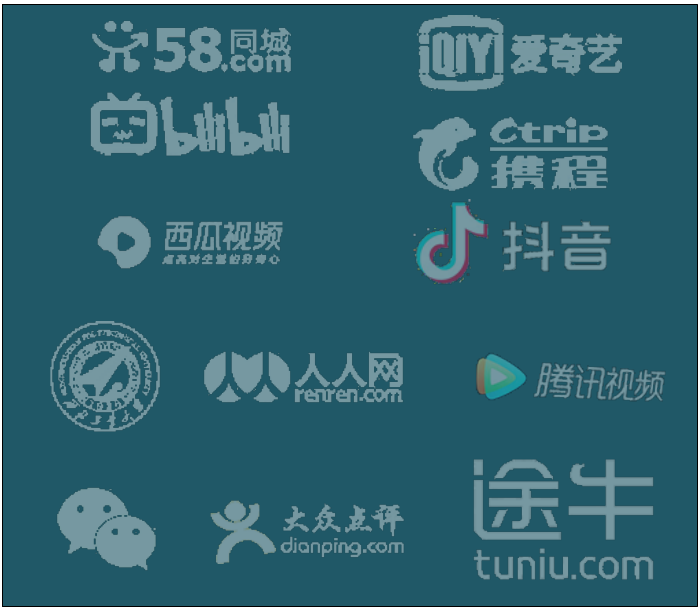}
  }
\caption{Twelve collected watermarks.}
 \vspace{-2.0em}
\end{figure}
\subsection{Experimental datasets}

Training datasets: 477 natural images from the PASCAL VOC 2021 \cite{everingham2015pascal} is chosen to conducted training datasets, where each image is saved in format of ‘.jpg’. Specifically, each watermarked image is randomly added one of twelve watermarks as shown in Fig. 2 with random one of four transparency, i.e., 0.3, 0.5, 0.7 and 1.0, coverage from 0 to 0.4, watermark size from 0.5 to 1 time, where twelve watermarks with different sizes and shapes are representative. Also, the diversity of these watermarks can enhance the robustness of a watermark removal model. To obtain watermarked images with noise, Gaussian noise of random one from noise levels, i.e., 0, 25, and 50 is added to the watermarked image. To improve training efficiency, each watermarked image with noise is cropped as 3,111 patches of $256\times256$.

Test dataset: Twenty-one natural images in the PASCAL VOV 2012 \cite{everingham2015pascal} are chosen to conduct the test dataset. Each watermark image with noise is conducted via randomly adding one watermark from twelve watermarks and fixed noise from noise levels of 15, 25, and 50. The number of test watermarked images is 252.

%%%%%%%%%%%%%%%%%%%%%%%%%%%%%%%%%%%%%%%Figure 2%%%%%%%%%%%%%%%%%%%%%%%%%%%%%%%%%%%%%%%%%%%%%%%%%%%%%%%%%%

\subsection{Experimental setting}
All the experiments can be run on a PC with Ubuntu of 20.04 and Intel Xeon Silver 
4210. Also, it depends on PyTorch \cite{NEURIPS2019_9015} of 1.8 and Python of 3.6 to implement codes. A GPU, i.e. NVIDIA RTX 3090 cooperates with CUDA 11.1, and CuDNN 8.0.5 to improve training speed. Besides, original training parameters are set to batch size of 8, epoch number of 100, and initial learning rate of 1e-3, where the original learning rate will vary 0.1 times every 30 times. More parameters set can be shown in Ref. \cite{zhang2017beyond}.

\subsection{Experimental analysis}

\begin{table}[]
\centering
\caption{Average PSNR (dB) and RMSE results of some methods with the noise level of 25 and watermark transparency of 0.3.}
\begin{tabular}{ccc}
\hline
Methods                                                                                                  & PSNR (dB) & RMSE   \\ \hline
Lower network with only a IUNet                                                                          & 28.9874  & 9.1154 \\
Lower network                                                                                            & 29.6631  & 8.5903 \\
Upper network                                                                                            & 29.7114  & 8.5306 \\
\begin{tabular}[c]{@{}c@{}}PSLNet without a combination of\\  Interaction1 and Interaction2\end{tabular} & \textcolor{blue}{29.7585}  & \textcolor{blue}{8.5130} \\
PSLNet (Ours)                                                                                             & \textcolor{red}{29.8154}  & \textcolor{red}{8.4557} \\ \hline
\end{tabular}

\end{table}

\begin{table}[]
\centering
\caption{Avergage PSNR (dB) and RMSE results of two methods with noise level of 25 and watermark transparency of 0.3.}
\begin{tabular}{ccc}
\hline
Methods                          & PSNR (dB)    & RMSE   \\ \hline
PSLNet with only structural loss & 28.7468 & 9.3163 \\
PSLNet (Ours)                           & 29.8154 & 8.4557 \\ \hline
                                 &         &       
\end{tabular}
\vspace{-2.0em}
\end{table}

\begin{table}[]
\centering

\caption{Average PSNR (dB) and RMSE of two methods with noise level of 25 and watermark transparency of 0.3.}

\begin{tabular}{ccc}
\hline
Methods        & PSNR (dB)    & RMSE   \\ \hline
PSLNet with L2 & 28.7105 & 9.4999 \\
PSLNet with L1 (Ours) & 29.8154 & 8.4557 \\ \hline
               &         &       
\end{tabular}

\end{table}

\begin{table}[]
\caption{Average PSNR (dB), SSIM and LPIPS of different methods for noise levels of 0, 15, 25 and 50 with watermark transparency of 0.3.}
\scalebox{0.8}{
\begin{tabular}{ccccccc}
\hline
Methods                & PSNR   & SSIM   & LPIPS  & PSNR   & SSIM   & LPIPS  \\ \hline
Noise levels           & \multicolumn{3}{c}{$\sigma$ = 0}  & \multicolumn{3}{c}{$\sigma$ = 15} \\ \hline
DnCNN         & 31.27  & 0.9482 & 0.0211 & 30.44  & 0.8833 & \textcolor{blue}{0.1455} \\
FFDNet        & 28.82  & 0.8904 & 0.1019 & 29.03  & 0.8570 & 0.1755 \\
IRCNN         & 32.21  & \textcolor{blue}{0.9824} & 0.0211 & \textcolor{blue}{29.57}  & \textcolor{blue}{0.8734} & 0.1591 \\
FastDerainNet & \textcolor{blue}{34.44}  & 0.9807 & \textcolor{blue}{0.0145} & 29.14  & 0.8550 & 0.1582 \\
DRDNet        & 31.97  & 0.9745 & 0.0305 & 27.24  & 0.8585 & 0.1706 \\
PSLNet (Ours)         & \textcolor{red}{42.16}  & \textcolor{red}{0.9932} & \textcolor{red}{0.0043} & \textcolor{red}{32.07}  & \textcolor{red}{0.8972} & \textcolor{red}{0.1320} \\ \hline
Noise levels           & \multicolumn{3}{c}{$\sigma$ = 25} & \multicolumn{3}{c}{$\sigma$ = 50} \\ \hline
DnCNN         & \textcolor{blue}{28.81}  & \textcolor{blue}{0.8231} & \textcolor{blue}{0.2163} & 25.64  & 0.6934 & 0.3406 \\
FFDNet        & 26.84  & 0.7888 & 0.2509 & 25.17  & 0.6959 & 0.3537 \\
IRCNN         & 27.67  & 0.8008 & 0.2406 & 24.91  & 0.6795 & 0.3642 \\
FastDerainNet & 26.25  & 0.7799 & 0.2364 & 24.85  & 0.6821 & 0.3430 \\
DRDNet        & 26.53  & 0.8104 & 0.2280 & \textcolor{blue}{25.83}  & \textcolor{blue}{0.7261} & \textcolor{blue}{0.3090} \\
PSLNet (Ours)         & \textcolor{red}{29.82}  & \textcolor{red}{0.8434} & \textcolor{red}{0.1959} & \textcolor{red}{26.90}  & \textcolor{red}{0.7499} & \textcolor{red}{0.2992} \\
\hline
\end{tabular}
}
\end{table}

\begin{table}[]
\caption{Average PSNR (dB), SSIM and LPIPS of different methods with noise level of 25 with blind watermark transparency of 0.3, 0.5, 0.7, and 1.0.}
\scalebox{0.8}{
\begin{tabular}{ccccccc}
\hline
Methods                & PSNR    & SSIM     & LPIPS    & PSNR    & SSIM     & LPIPS    \\ \hline
Transparency          & \multicolumn{3}{c}{Alpha = 0.3} & \multicolumn{3}{c}{Alpha = 0.5} \\ \hline
DnCNN         & 27.01   & 0.8008   & \textcolor{blue}{0.2333}   & 26.95   & 0.8006   & \textcolor{blue}{0.2339}   \\
FFDNet        & 25.34   & 0.7630   & 0.2781   & 23.97   & 0.7570   & 0.2834   \\
IRCNN         & \textcolor{blue}{26.30}   & \textcolor{blue}{0.8084}   & 0.2392   & \textcolor{blue}{26.13}   & \textcolor{blue}{0.8064}   & 0.2415   \\
FastDerainNet & 25.89   & 0.7724   & 0.2456   & 25.90   & 0.7720   & 0.2463   \\
DRDNet        & 24.01   & 0.7720   & 0.2630   & 24.51   & 0.7731   & 0.2625   \\
PSLNet (Ours)          & \textcolor{red}{28.43}   & \textcolor{red}{0.8335}   & \textcolor{red}{0.2078}   & \textcolor{red}{28.01}   & \textcolor{red}{0.8311}   & \textcolor{red}{0.2104}   \\ \hline
Transparency          & \multicolumn{3}{c}{Alpha = 0.7} & \multicolumn{3}{c}{Alpha = 1.0} \\ \hline
DnCNN         & 27.65   & 0.8021   & \textcolor{blue}{0.2332}   & 21.40   & 0.7878   & 0.2457   \\
FFDNet        & 22.86   & 0.7545   & 0.2856   & 25.30   & 0.7623   & 0.2799   \\
IRCNN         & \textcolor{blue}{25.87}   & \textcolor{blue}{0.8040}   & 0.2436   & \textcolor{blue}{25.62}   & \textcolor{blue}{0.8011}   & 0.2458   \\
FastDerainNet & 25.76   & 0.7701   & 0.2484   & 21.17   & 0.7585   & 0.2589   \\
DRDNet        & 24.49   & 0.7704   & 0.2650   & 20.61   & 0.7605   & 0.2735   \\
PSLNet (Ours)         & \textcolor{red}{27.87}   & \textcolor{red}{0.8310}   & \textcolor{red}{0.2105}   & \textcolor{red}{28.03}   & \textcolor{red}{0.8329}   & \textcolor{red}{0.2088}   \\ \hline
\end{tabular}
}
\end{table}

\begin{table}[]
\caption{Average PSNR (dB), SSIM and LPIPS of different methods fixed watermark transparency of 0.3 with noise level of 0, 15, 25, 50. }
\scalebox{0.8}{
\begin{tabular}{ccccccc}
\hline
Methods                & PSNR   & SSIM   & LPIPS  & PSNR   & SSIM   & LPIPS  \\ \hline
Noise levels           & \multicolumn{3}{c}{$\sigma$ = 0}  & \multicolumn{3}{c}{$\sigma$ = 15} \\ \hline
DnCNN         & \textcolor{blue}{35.13}  & \textcolor{blue}{0.9794} & \textcolor{red}{0.0205} & \textcolor{blue}{29.86}  & \textcolor{blue}{0.8652} & \textcolor{blue}{0.1648} \\
FFDNet        & 27.39  & 0.8564 & 0.1548 & 26.91  & 0.8048 & 0.2224 \\
IRCNN         & 32.61  & 0.9684 & 0.0335 & 29.10  & 0.8624 & 0.1747 \\
FastDerainNet & 29.85  & 0.9336 & 0.0714 & 27.88  & 0.8352 & 0.1863 \\
DRDNet        & 31.56  & 0.9516 & 0.0517 & 29.40  & 0.8578 & 0.1699 \\
PSLNet (Ours)         & \textcolor{red}{35.55}  & \textcolor{red}{0.9732} & \textcolor{blue}{0.0273} & \textcolor{red}{30.99}  & \textcolor{red}{0.8866} & \textcolor{red}{0.1433} \\ \hline
Noise levels           & \multicolumn{3}{c}{$\sigma$ = 25} & \multicolumn{3}{c}{$\sigma$ = 50} \\ \hline
DnCNN         & \textcolor{blue}{27.87}  & 0.7951 & 0.2379 & \textcolor{blue}{24.73}  & 0.6449 & 0.3645 \\
FFDNet        & 26.27  & 0.7648 & 0.2693 & 24.69  & 0.6778 & 0.3701 \\
IRCNN         & 27.60  & \textcolor{blue}{0.8032} & 0.2416 & 25.28  & \textcolor{blue}{0.6917} & 0.3600 \\
FastDerainNet & 26.70  & 0.7760 & 0.2480 & 24.59  & 0.6474 & 0.3598 \\
DRDNet        & 27.65  & 0.7908 & \textcolor{blue}{0.2375} & 24.63  & 0.6529 & \textcolor{blue}{0.3545} \\
PSLNet (Ours)         & \textcolor{red}{29.13}  & \textcolor{red}{0.8346} & \textcolor{red}{0.2059} & \textcolor{red}{26.44}  & \textcolor{red}{0.7361} & \textcolor{red}{0.3124} \\ \hline
\end{tabular}
}
\end{table}

\begin{table}[]
\caption{Average PSNR (dB), SSIM and LPIPS of different methods for blind noise levels with watermark transparency of 0.3.}
\scalebox{0.8}{
\begin{tabular}{ccccccc}
\hline
Methods                & PSNR   & SSIM   & LPIPS  & PSNR   & SSIM   & LPIPS  \\ \hline
Noise levels           & \multicolumn{3}{c}{$\sigma$ = 0}  & \multicolumn{3}{c}{$\sigma$ = 15} \\ \hline
DnCNN         & 28.65  & 0.9590 & 0.0436 & 26.53  & 0.8413 & 0.1855 \\
FFDNet        & 26.04  & 0.8173 & 0.1959 & 25.71  & 0.7829 & 0.2423 \\
IRCNN         & \textcolor{blue}{29.28}  & \textcolor{blue}{0.9635} & \textcolor{blue}{0.0386} & \textcolor{blue}{26.94}  & \textcolor{blue}{0.8569} & \textcolor{blue}{0.1834} \\
FastDerainNet & 27.24  & 0.9134 & 0.0680 & 26.17  & 0.8174 & 0.1845 \\
DRDNet        & 24.27  & 0.8873 & 0.1108 & 23.10  & 0.7935 & 0.2270 \\
PSLNet (Ours)         & \textcolor{red}{35.93}  & \textcolor{red}{0.9777} & \textcolor{red}{0.0176} & \textcolor{red}{31.09}  & \textcolor{red}{0.8887} & \textcolor{red}{0.1419} \\ \hline
Noise levels           & \multicolumn{3}{c}{$\sigma$ = 25} & \multicolumn{3}{c}{$\sigma$ = 50} \\ \hline
DnCNN         & 25.22  & 0.7624 & 0.2599 & 22.43  & 0.5913 & 0.3955 \\
FFDNet        & 25.25  & 0.7511 & 0.2825 & \textcolor{blue}{24.02}  & \textcolor{blue}{0.6781} & 0.3699 \\
IRCNN         & \textcolor{blue}{25.70}  & \textcolor{blue}{0.7977} & 0.2528 & 23.45  & 0.6779 & 0.3837 \\
FastDerainNet & 25.23  & 0.7542 & \textcolor{blue}{0.2491} & 23.21  & 0.6143 & \textcolor{blue}{0.3695} \\
DRDNet        & 21.59  & 0.7158 & 0.2987 & 19.02  & 0.5681 & 0.4260 \\
PSLNet (Ours)         & \textcolor{red}{29.27}  & \textcolor{red}{0.8382} & \textcolor{red}{0.2041} & \textcolor{red}{26.58}  & \textcolor{red}{0.7405} & \textcolor{red}{0.3143} \\ \hline
\end{tabular}
}
\vspace{-2.0em}
\end{table}

%%%%%%%%%%%%%%%%%%%%%%%%%%%%%%Table %%

\begin{table*}[]
\centering
\caption{Average PSNR (dB), SSIM, and LPIPS of different methods trained on blind noise level and blind watermark transparency and tested with the fixed noise level of 25 and certain watermark transparency of 0.5, 0.7, and 1.0.}

\begin{tabular}{cccccccccc}
\hline
Methods                & PSNR    & SSIM     & LPIPS    & PSNR    & SSIM     & LPIPS    & PSNR    & SSIM     & LPIPS    \\ \hline
Transparency          & \multicolumn{3}{c}{Alpha = 0.5} & \multicolumn{3}{c}{Alpha = 0.7} & \multicolumn{3}{c}{Alpha = 1.0} \\ \hline
DnCNN         & 23.88   & 0.7567   & 0.2649   & 23.22   & 0.7555   & 0.2659   & \textcolor{blue}{25.29}   & 0.7601   & 0.2622   \\
FFDNet        & 25.27   & 0.7514   & 0.2832   & 25.23   & 0.7501   & 0.2848   & 21.20   & 0.7379   & 0.2944   \\
IRCNN         & \textcolor{blue}{25.52}   & \textcolor{blue}{0.7951}   & 0.2557   & \textcolor{blue}{25.29}   & \textcolor{blue}{0.7927}   & 0.2578   & 24.96   & \textcolor{blue}{0.7900}   & \textcolor{blue}{0.2599}   \\
FastDerainNet & 25.27   & 0.7529   & \textcolor{blue}{0.2505}   & 25.07   & 0.7505   & \textcolor{blue}{0.2530}   & 20.76   & 0.7397   & 0.2633   \\
DRDNet        & 21.52   & 0.7130   & 0.3026   & 21.77   & 0.7149   & 0.3002   & 18.25   & 0.7064   & 0.3085   \\
PSLNet (Ours)         & \textcolor{red}{29.05}   & \textcolor{red}{0.8364}   & \textcolor{red}{0.2061}   & \textcolor{red}{28.61}   & \textcolor{red}{0.8341}   & \textcolor{red}{0.2084}   & \textcolor{red}{27.90}   & \textcolor{red}{0.8308}   & \textcolor{red}{0.2116}   \\ \hline
\end{tabular}

\end{table*}

\begin{table*}[]
\centering
\caption{Average PSNR (dB), SSIM and LPIPS of different methods trained with blind noise and without watermarks, and tested with certain noise levels of 15, 25, 50 without watermarks.}

\begin{tabular}{cccccccccc}
\hline
Methods                & PSNR   & SSIM   & LPIPS  & PSNR   & SSIM   & LPIPS  & PSNR   & SSIM   & LPIPS  \\ \hline
Noise levels           & \multicolumn{3}{c}{$\sigma$ = 15} & \multicolumn{3}{c}{$\sigma$ = 25} & \multicolumn{3}{c}{$\sigma$ = 50} \\ \hline
DnCNN         & 30.42  & 0.8715 & \textcolor{blue}{0.1581} & 28.26  & 0.8011 & \textcolor{blue}{0.2319} & 24.94  & 0.6501 & 0.3599 \\
FFDNet        & 27.50  & 0.8125 & 0.2142 & 26.76  & 0.7716 & 0.2624 & 25.03  & \textcolor{blue}{0.6832} & 0.3655 \\
IRCNN         & \textcolor{blue}{30.78}  & \textcolor{blue}{0.8737} & 0.1634 & \textcolor{blue}{28.73}  & \textcolor{blue}{0.8131} & 0.2325 & \textcolor{blue}{25.95}  & 0.6695 & 0.3543 \\
FastDerainNet & 28.53  & 0.8421 & 0.1792 & 27.24  & 0.7825 & 0.2418 & 24.98  & 0.6531 & 0.3550 \\
DRDNet        & 29.81  & 0.8624 & 0.1654 & 27.92  & 0.7948 & 0.2337 & 24.81  & 0.6564 & \textcolor{blue}{0.3512} \\
PSLNet (Ours)         & \textcolor{red}{31.68}  & \textcolor{red}{0.8919} & \textcolor{red}{0.1379} & \textcolor{red}{29.58}  & \textcolor{red}{0.8395} & \textcolor{red}{0.2012} & \textcolor{red}{26.72}  & \textcolor{red}{0.7406} & \textcolor{red}{0.3082} \\ \hline
\end{tabular}

\end{table*}

\begin{table}[]
\caption{Average PSNR (dB), SSIM and LPIPS of different methods trained blind watermark transparency without noise and tested certain watermark transparency of 0.3, 0.5. 0.7, 1.0 without noise.}
\scalebox{0.8}{
\begin{tabular}{ccccccc}
\hline
Methods                & PSNR    & SSIM     & LPIPS    & PSNR    & SSIM     & LPIPS    \\ \hline
Transparency          & \multicolumn{3}{c}{Alpha = 0.3} & \multicolumn{3}{c}{Alpha = 0.5} \\ \hline
DnCNN         & 29.49   & 0.9406   & 0.0617   & 29.39   & 0.9405   & 0.0614   \\
FFDNet        & 25.87   & 0.8548   & 0.1410   & 25.83   & 0.8551   & 0.1408   \\
IRCNN         & \textcolor{blue}{31.21}   & 0.9673   & 0.0264   & 31.12   & 0.9659   & 0.0279   \\
FastDerainNet & 26.97   & 0.9508   & 0.0274   & 26.58   & 0.9504   & \textcolor{blue}{0.0263}   \\
DRDNet        & 31.02   & \textcolor{blue}{0.9763}   & \textcolor{blue}{0.0267}   & \textcolor{blue}{31.26}   & \textcolor{blue}{0.9752}   & 0.0284   \\
PSLNet (Ours)         & \textcolor{red}{38.66}   & \textcolor{red}{0.9909}   & \textcolor{red}{0.0075}   & \textcolor{red}{38.48}   & \textcolor{red}{0.9903}   & \textcolor{red}{0.0081}   \\ \hline
Transparency          & \multicolumn{3}{c}{Alpha = 0.7} & \multicolumn{3}{c}{Alpha = 1.0} \\ \hline
DnCNN         & 29.21   & 0.9389   & 0.0633   & 22.22   & 0.9211   & 0.0819   \\
FFDNet        & 25.83   & 0.8536   & 0.1425   & 21.43   & 0.8406   & 0.1546   \\
IRCNN         & \textcolor{blue}{30.95}   & 0.9637   & 0.0306   & \textcolor{blue}{29.71}   & 0.9585   & \textcolor{blue}{0.0361}   \\
FastDerainNet & 26.11   & 0.9479   & \textcolor{blue}{0.0285}   & 21.27   & 0.9358   & 0.0433   \\
DRDNet        & 29.92   & \textcolor{blue}{0.9709}   & 0.0330   & 22.91   & \textcolor{blue}{0.9605}   & 0.0430   \\
PSLNet (Ours)         & \textcolor{red}{37.40}   & \textcolor{red}{0.9884}   & \textcolor{red}{0.0103}   & \textcolor{red}{34.50}   & \textcolor{red}{0.9820}   & \textcolor{red}{0.0177}   \\ \hline
\end{tabular}
}

\end{table}

\begin{table}[]
\caption{Complexity of different methods on an image with 256$\times$256 for noisy image watermark removal.}
\centering
\begin{tabular}{lll}
\cline{1-3}
Methods        & Parameters & FLOPs      \\ \cline{1-3}
DnCNN  & 0.558M   & 36.591G   \\
DRDNet & 2.941M     & 192.487G \\
PSLNet         & 2.516M     & 74.513G   \\ \cline{1-3}
\end{tabular}
\vspace{-2.0em}
\end{table}

Watermark images can protect the copyright of interactive images \cite{wong2003novel}. To verify the robustness of obtained watermarks, watermark removal techniques are proposed \cite{fu2022improved}. Existing watermark removal techniques use supervised methods to remove watermarks. However, reference images are not easy to obtain. Additionally, collected images have noise in general. To overcome the mentioned two challenges, we propose a perceptive self-supervised learning network for noisy image watermark removal. That mainly includes three phases. The first phase uses a self-supervised mechanism to obtain paired training images, i.e., given watermark image with noise and given watermark image, according to noise-to-noise \cite{lehtinen2018noise2noise}. Its rational analysis has been given in Section II.A. The second and third phases can be guided via structural and texture information.   

The second phase guides a CNN for noisy image watermark removal, according to complex task and structural information. In terms of complex task (images are simultaneously damaged by factors, i.e., camera shake, long exposure), i.e., noisy image super-resolution and noisy image deblurring, scholars use a degradation model to simultaneously suppress multiple damaged factors for recovering high-quality images \cite{tian2021asymmetric}. For instance, Tian et al. \cite{tian2021asymmetric} use an asymmetric architecture to extract salient features in terms of horizontal and vertical for simultaneously removing watermarks and noise. Alternatively, task decomposition is also an effective tool for complex tasks \cite{guo2023joint}\cite{tan2022two}. For instance, Guo et al. used a traditional color filter array processing pipeline to remove mosaicking and then exploited a CNN to filter noise \cite{guo2023joint}.

Although these methods can better address complex tasks in image restoration, their performance can be further improved as follows. (1) The first method may ignore the relationship between each task. (2) The second method uses two phases to finish different low-level vision tasks in sequence for image restoration, which may loss of key information. Taking into the mentioned illustrations account, we propose a parallel network for noisy image watermark removal. Upper and lower networks are composed of two improved U-Nets, which can be shown in Section II.B. Also, the upper network is used to first remove noise and then remove watermarks. The lower network is used to simultaneously filter noise and watermarks. To enhance the clarity of obtained images, multiple interactions are conducted to act between upper and lower networks for image watermark removal. The first interaction is Interaction1, which is applied on the first improved U-Nets of upper and lower networks to filter noise. The second interaction is Interaction2, which is applied to the second improved U-Nets of upper and lower networks to suppress watermarks. Interaction1 and Interaction2 are introduced in Section II.A as shown in Fig. 1. The first and second interactions can improve structural information. As shown in TABLE I, we can see that our PSLNet has obtained a higher PSNR \cite{hore2010image} value than that of PSLNet without a combination of Interaction1 and Interaction2. Also, PSLNet has a lower value of RMSE \cite{chai2014root} than that of PSLNet without a combination of Interaction 1 and Interaction2 in TABLE I. That shows the effectiveness of a combination of Interaction1 and Interaction2 for noisy image watermark removal. The third interaction can use an EM composed of a concatenation operation, a convolutional layer, and a LeakyReLU to fuse two obtained images from upper and lower networks to enhance obtained final images, according to pixel enhancement. The third interaction is a basic interaction between upper and lower networks. Besides, PSLNet has better PSNR and RMSE values than that of a lower network and an upper network in TABLE I, which shows the effectiveness of a lower network and an upper network in the PSLNet for noisy image watermark removal. The lower network has an improvement of 0.6757dB for PSNR and 0.5251 for RMSM than that of an IUNet in TABLE I, which shows the effectiveness of two stacked IUNets in the lower network for noisy image watermark removal. Besides, red and blue lines denote the best and second performance from TABLE I, respectively.

Taking texture information into account, the third phase uses the perceptual network to extract features to compute texture loss value, as shown in Section II.B, which shows the effectiveness of mixed loss via comparing PSLNet and PSLNet with only structural loss in terms of PSNR and RMSE in TBALE II. Taking training speed into account, L1 loss is chosen. As shown in TABLE III, we can see that PSLNet with L1 has obtained better performance than that of PSLNet with L2 for PSNR and RMSE, which verifies the effectiveness of L1 in the PSLNet for noisy image watermark removal. This analysis shows the effectiveness and rationality of our PSLNet for noisy image watermark removal. 

\subsection{Experimental results}
\begin{figure*}[!htbp]
\centering
\subfloat{\includegraphics[width=7in]{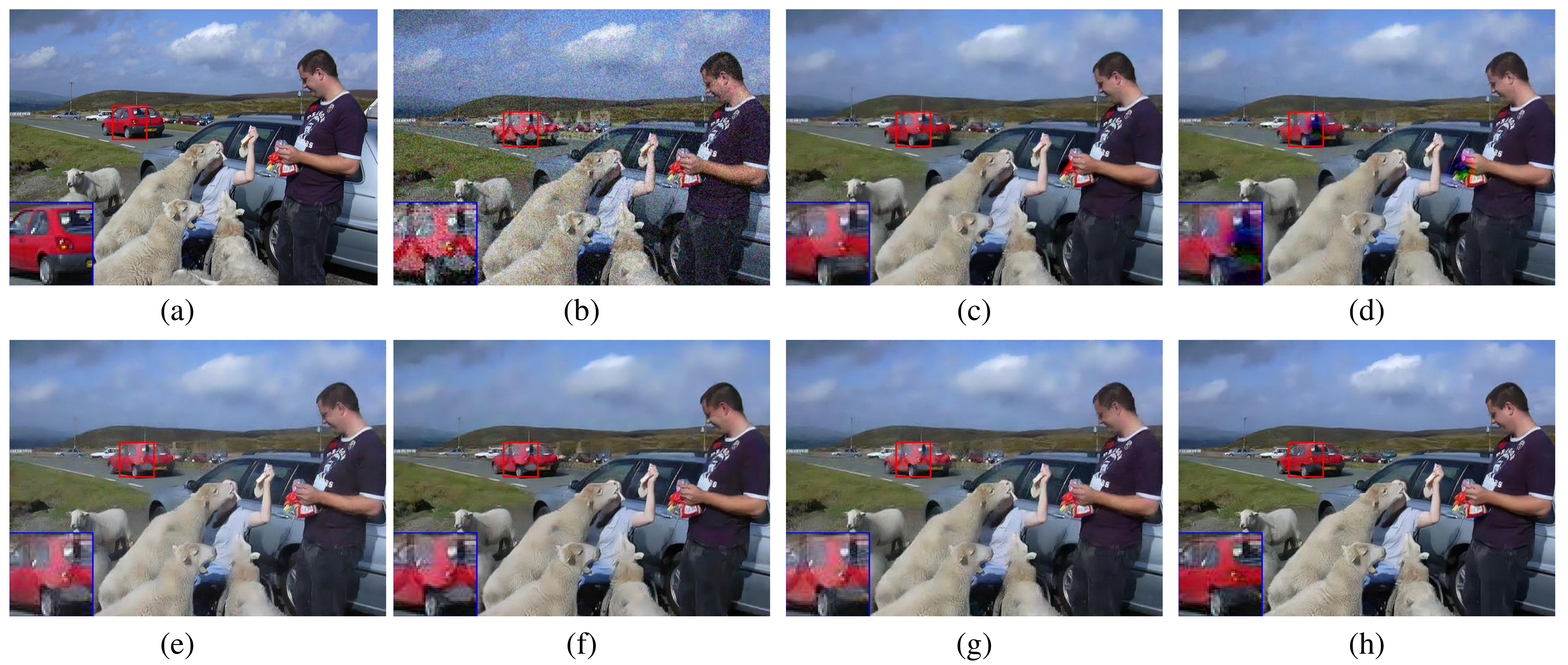}
  }
\caption{ Results of different methods on one image from test dataset when $\sigma$ = 25 and transparency = 0.3. (a) Original image (b) Noisy image/20.02 dB (c) DnCNN/28.50 dB (d) DRDNet/27.03 dB (e) FastDerainNet/.26.32 dB (f) FFDNet/26.98 dB (g) IRCNN/27.39 dB (h) PSLNet/29.72 dB.} 
\vspace{-1.0em}
\end{figure*}

 \begin{figure*}[!htbp]
\centering
\subfloat{\includegraphics[width=7in]{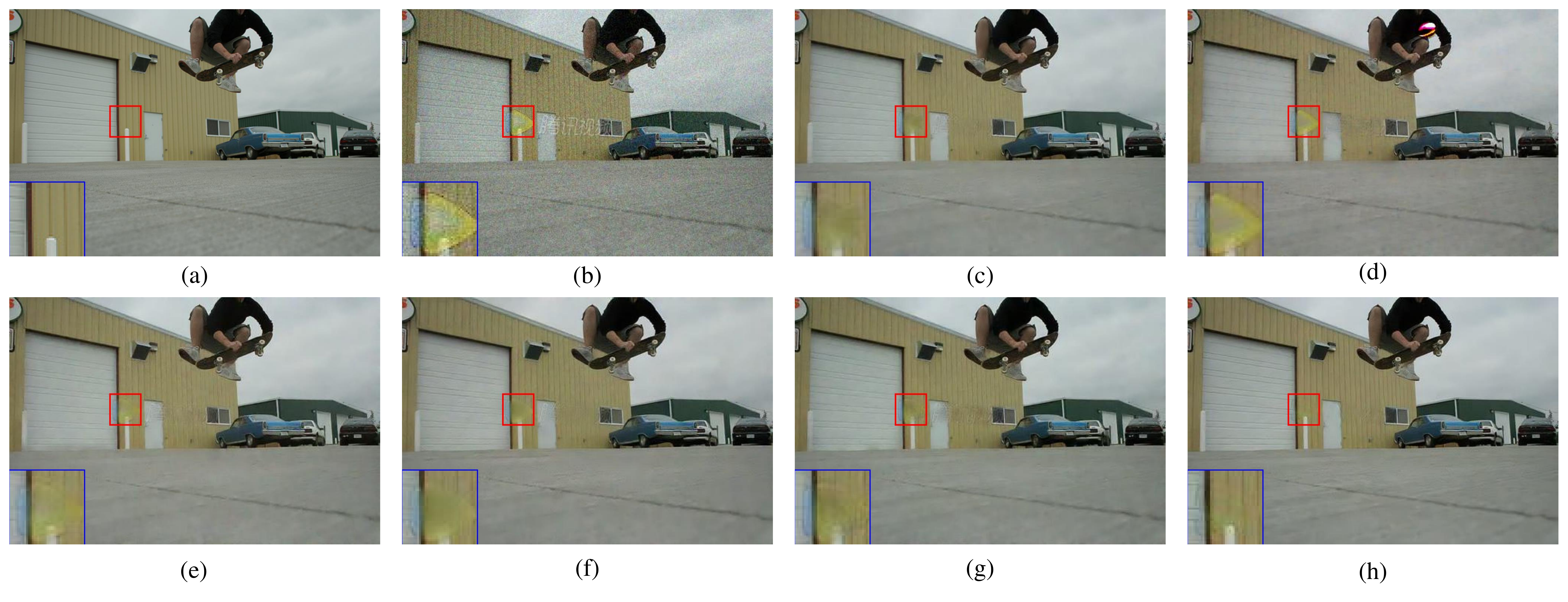}
  }
\caption{ Results of different methods on one image from test dataset when $\sigma$ = 15 and transparency = 0.3. (a) Original image (b) Noisy image/24.42 dB (c) DnCNN/34.15 dB (d) DRDNet/27.46 dB (e) FastDerainNet/31.88 dB (f) FFDNet/32.67 dB (g) IRCNN/32.96 dB (h) PSLNet/35.19 dB.} 

\end{figure*}
We evaluate the noise and watermark removal performance of our proposed method by conducting quantitative and qualitative analysis. Serval popular restoration methods, i.e., denoising CNN (DnCNN) \cite{zhang2017beyond}, fast and flexible denoising CNN (FFDNet)\cite{zhang2018ffdnet}, an image restoration CNN (IRCNN) \cite{zhang2017learning}, a deep residual learning algorithm for removing rain streaks (FastDerainNet)\cite{wang2020fastderainnet} and detail-recovery image draining network (DRDNet)\cite{deng2019drd} are conducted as comparative methods for noisy image removal. 

Quantitative analysis uses different methods with different conditions, i.e.,  transparency and noise levels
to obtain PSNR\cite{hore2010image}, SSIM\cite{wang2004image}, Learned Perceptual Image Patch Similarity (LPIPS) \cite{zhang2018unreasonable} for testing performance of noisy watermark removal. To further test the robustness of PSLNet, we conduct serval experiments, i.e., watermark removal of specific transparency with specific noise removal, watermark removal of blind transparency with specific noise level, watermark removal of specific transparency watermark with blind noise, watermark removal of blind transparency with blind noise, blind image watermark removal without noise, blind image denoising without watermarks and complexity. For watermark removal of specific transparency with specific noise removal, we choose noise levels of 0, 15, 25, and 50, and fixed watermark transparency of 0.3 to train and test our watermark removal model with noise, respectively. As shown in TABLE IV, we can see that our PSLNet has obtained better results than that of popular DnCNN, IRCNN, and FastDerainNet for different noise levels, i.e., 0, 15, 25 and 50 watermark transparency of 0. 3 in terms of PSNR, SSIM, and LPISP \cite{zhang2018unreasonable}. For watermark removal of blind transparency with a specific noise level, we choose a noise level of 25 and blind watermark transparency, which varies from 0.3 to 1.0 when our PSLNet is trained. As reported in TABLE V, we can see that our PSLNet is more effective than those of the compared methods, i.e., IRCNN and FastDerainNet. For instance, our PSLNet has an improvement of 2.11dB in terms of PSNR for a noise level of 25 with a watermark transparency of 0.3 than that of the second IRCNN. For watermark removal of specific transparency watermarks with blind noise, a certain watermark transparency of 0.3 and varying noise from 0 to 55 are used to conduct comparative experiments in the training process. Also, certain noise of 0, 15, 25, and 50 with a certain watermark transparency of 0.3 is used to test the performance of our PSLNet for noisy image removal. As shown in TABLE VI, our PSLNet is superior to other comparative methods, which shows the validity of PSLNet for image denoising and watermark removal. For instance, our PSLNet has an improvement of 0.0444 in terms of SSIM than that of the second IRCNN for noise of 0.5 with watermark transparency of 0.3. For image watermark removal of blind transparency with blind noise, our PSLNet has the best performance as shown in TABLEs VII and VIII. TABLE VII shows the superiority of our method for blind noise levels from 0 to 55 with blind watermark transparency from 0.3 to 1 in the training process and certain noise levels from 0, 15, 25, and 50 with fixed watermark transparency of 0.3 in the test process. TABLE VIII illustrates the effectiveness of our method for blind noise levels from 0 to 55 with blind watermark transparency from 0.3 to 1 in the training process and a certain noise level of 25 with fixed watermark transparency of 0.5, 0.7, and 1 in the test process. To verify the effectiveness of our method for only removing watermarks without noise, we choose varying watermark transparency from 0.3 to 1 to train a single watermark removal model. Also, watermark transparency is set to single 0.3, 0.5, and 0.7 to test our PSLNet for image watermark removal. As shown in TABLE X, we can see that our PSLNet still obtains the best performance for different watermark transparency, which shows its effectiveness for image watermark removal. To verify the effectiveness of our method for only removing noise without watermarks, we choose varying noise levels from 0 to 55 to train a single watermark removal model. Also, the noise level is set to single 15, 25, and 55 to test our PSLNet for image denoising. As shown in TABLE IX, we can see that our PSLNet is more effective than other methods for all noise levels. Besides, to verify practicality of our PSLNet for digital devices, we use parameters and flops to test complexity of PSLNet. As shown in TABLE XI, our PSLNet has less parameters than that DnCNN and DRDNet. Although PSLNet has higher flops than that of DnCNN, our watermark removal performance is much higher than DnCNN for different noise. Besides, the red and blue lines denote the best and second performance from TABLE IV to TABLE X, respectively. Thus, our PSLNet is very competitive for complexity. According to the mentioned illustrations, we can see that our PSLNet is very effective in terms of quantitative analysis.

In terms of qualitative analysis, popular methods, i.e., DnCNN, DRDNet, FastDerainNet, and FFDNet are selected to obtain visually predicted images, and an area of these images is amplified as an observation area. If the observation area is clearer, its corresponding method has better performance. As shown in Figs. 3 and 4, we can see that observation images from our PSLNet are clearer than other methods for noise level 25 and 15 with transparency of 0.3, respectively. That demonstrates that our method has better visual effects. According to quantitative and qualitative analysis, it is known that our PSLNet is suitable for watermark removal with noise. 

\section{Conclusion}
In this paper, we have presented a perceptive self-supervised learning network for noisy image watermark removal i.e. PSLNet. PSLNet depends on a parallel network to remove noise and watermarks, according to solution ideas of the inverse problem. The upper network is used to successively remove noise and watermarks, according to the task decomposition idea. The lower network is used to simultaneously remove noise and watermarks, according to the degradation model idea. To address the problem of difficulty obtaining reference images, a self-supervised learning method is used to obtain reference images, according to given watermarked images. To obtain clearer images, two sub-networks and their obtained images are respectively integrated to enhance structural information and pixels. Finally, a mixed loss uses obtained images and features to facilitate more texture information, according to perceptual ideas and pixel relations. Comprehensive evaluations show that our PSLNet is very effective for noisy image watermark removal. In the future, we will combine lightweight networks and self-supervised techniques in image watermark removal.

\ifCLASSOPTIONcaptionsoff
  \newpage
\fi

\bibliographystyle{IEEEtran}
\bibliography{IMSC_AGL}

\end{document}